\documentclass{article}

\usepackage{times}
\usepackage[utf8]{inputenc} %
\usepackage[T1]{fontenc}    %
\usepackage{url}            %
\usepackage{booktabs}       %
\usepackage{nicefrac}       %
\usepackage{microtype}      %
\usepackage{graphics,color}

\usepackage[ruled,vlined,linesnumbered]{algorithm2e}
\DontPrintSemicolon
\SetAlgoLined
\usepackage{enumitem}

\usepackage{amsfonts}       %
\usepackage{mathtools}
\usepackage{amssymb}

\newcommand{\Fig}[1]{Figure~\ref{#1}}  %
\newcommand{\fig}[1]{Fig.~\ref{#1}}    %
\newcommand{\tab}[1]{Table~\ref{#1}}

\newcommand{\eqn}[1]{Eq.~\ref{#1}} %
\renewcommand{\sec}[1]{Sec.~\ref{#1}} %
\newcommand{\supp}[1]{Suppl.~\ref{#1}}

\usepackage{xspace}
\makeatletter
\DeclareRobustCommand\onedot{\futurelet\@let@token\@onedot}
\def\@onedot{\ifx\@let@token.\else.\null\fi\xspace}
\def\eg{e.g\onedot}
\def\ie{i.e\onedot}

\makeatother

\usepackage{xcolor}
\definecolor{ourblue}{rgb}{0.368,0.507,0.71}
\definecolor{ourorange}{rgb}{0.881,0.611,0.142}
\definecolor{ourgreen}{rgb}{0.56,0.692,0.195}
\definecolor{ourred}{rgb}{0.923,0.386,0.209}
\definecolor{ourviolet}{rgb}{0.528,0.471,0.701}
\definecolor{ourbrown}{rgb}{0.772,0.432,0.102}
\definecolor{ourlightblue}{rgb}{0.364,0.619,0.782}
\definecolor{ourdarkgreen}{rgb}{0.572,0.586,0.}

\definecolor{ourcyan2}{rgb}{0.125,0.722,0.804}
\definecolor{ourred2}{rgb}{0.863,0.184,0.047}
\definecolor{ouryellow2}{cmyk}{0,0.16,1.0,0.07}
\definecolor{ourviolet2}{cmyk}{0.55,0.56,0,0.47}
\definecolor{ourorange2}{cmyk}{0,0.46,0.89,0.11}

\usepackage{amsmath,amsfonts,bm}

\def\1{\bm{1}}

\def\vmu{{\bm{\mu}}}

\def\vvartheta{{\bm{\vartheta}}}

\def\vu{{\bm{u}}}

\def\gH{{\mathcal{H}}}
\def\gI{{\mathcal{I}}}

\def\gL{{\mathcal{L}}}

\def\gN{{\mathcal{N}}}

\def\gU{{\mathcal{U}}}

\def\gX{{\mathcal{X}}}

\def\sC{{\mathbb{C}}}

\def\sR{{\mathbb{R}}}

\def\sX{{\mathbb{X}}}

\newcommand{\Var}{\mathrm{Var}}

\DeclareMathOperator{\E}{\mathbb{E}}

\usepackage{xr-hyper}

\makeatletter
\newcommand*{\addFileDependency}[1]{%
  \typeout{(#1)}
  \@addtofilelist{#1}
  \IfFileExists{#1}{}{\typeout{No file #1.}}
}
\makeatother

\usepackage{authblk}

\usepackage[preprint, nonatbib]{ewrl_2023} %
\usepackage{natbib}
\graphicspath{{../graphics/}}

\usepackage{caption}
\usepackage{subcaption}
\usepackage{mathtools}
\usepackage{amsthm}
\usepackage{stmaryrd}
\usepackage{wrapfig}
\usepackage[symbol]{footmisc}

\SetCommentSty{mycommfont}

\renewcommand{\paragraph}[1]{\par \textbf{#1.}~}
\usepackage[normalem]{ulem}
\usepackage[disable]{todonotes}

\usepackage{hyperref}
\hypersetup{
	colorlinks=true,       %
	linkcolor=ourblue,        %
	citecolor=ourblue,        %
	filecolor=ourblue,     %
	urlcolor=ourblue         
}

\newcommand{\method}{RAZER} %

\def\gp{{\vvartheta}}

\def\useq{{\vu}} %
\def\useqinit{{\useq_{t:t+h}}}
\def\px{{\tilde x}}
\def\mux{{\bar x}}
\def\muk{\mu^k_\theta}
\def\sigmak{\Sigma^k_\theta}
\def\ue{{\mathfrak{E}}}
\def\ua{{\mathfrak{A}}}
\def\sav{{\mathfrak{S}}}
\def\varale{\ensuremath{\mathrm{Var}^\ua}}
\def\varepi{\ensuremath{\mathrm{Var}^\ue}}
\def\xinit{{x_{t}}}
\def\trajdist{{\psi^\tau}}

\def\xtdist{{\psi_{\Delta t}^x}}
\def\xtbdist{{\psi_{\Delta t,b}^{x}}}

\def\gptdist{{\psi_{\Delta t}^\gp}}

\usepackage{listings}
\usepackage{multirow}

\usepackage[export]{adjustbox}
\usepackage{tabularx}
\usepackage{collcell}

\newcommand\clearrow{\global\let\rowmac\relax}
\clearrow

\newcolumntype{C}{>{\collectcell\rowmac}c<{\endcollectcell}}
\newcolumntype{R}{>{\collectcell\rowmac}r<{\endcollectcell}}
\newcolumntype{L}{>{\collectcell\rowmac}l<{\endcollectcell}}

\usepackage{hyperref}%
\title{Mind the Uncertainty: Risk-Aware and Actively Exploring Model-Based Reinforcement Learning}

\author[*,1]{\textbf{Marin Vlastelica}}
\author[*,1]{\textbf{Sebastian Blaes}}
\author[1]{\textbf{Cristina Pineri}}
\author[1,2]{\textbf{Georg Martius}}
\affil[1]{Max Planck Institute for Intelligent Systems, Tübingen, Germany}
\affil[2]{University of Tübingen, Tübingen, Germany}

\renewcommand{\cite}{\citep}

\begin{document}

\maketitle

\footnotetext[1]{Authors contributed equally.}
\footnotetext[2]{Published under the title \textit{Risk-Averse Zero-Order Trajctory Optimization} in the  Proceedings of the 5th Conference on Robot Learning, PMLR 164:444-454, 2022.}
\begin{abstract}
    We introduce a simple but effective method for managing risk in model-based reinforcement learning with trajectory sampling that involves probabilistic safety constraints and balancing of optimism in the face of epistemic uncertainty and pessimism in the face of aleatoric uncertainty of an ensemble of stochastic neural networks.
    Various experiments indicate that the separation of uncertainties is essential  to performing well with data-driven MPC approaches in uncertain and safety-critical control environments.
\end{abstract}

\section{Introduction}\label{sec:intro}

Data-driven approaches to sequential decision-making are becoming increasingly popular~\citep{yang2019Data, hussein2017imitation, polydoros2017survey, schrittwieser2020mastering}.
They hold the promise of reducing the number of prior assumptions about the system that are imposed by traditional approaches that are based on nominal models.

Such approaches come in several different flavors~\citep{kober2013reinforcement}.
Model-free approaches attempt to extract closed-loop control policies directly from data, while model-based approaches rely on a learned model of the dynamics to either generate novel data to extract a policy or to be used in a model-predictive control fashion (MPC). This study belongs to the latter line of work.

Model-based methods have several advantages over pure model-free approaches.
Firstly, humans tend to have a better intuition on how to incorporate prior knowledge into a model rather than into a policy or value function.
Secondly, most model-free policies are bounded to a specific task, while models are task-agnostic and can be applied for optimizing arbitrary cost functions, given sufficient exploration.

Nevertheless, learning models for control come with certain caveats.
Traditional MPC methods require the model and cost function to permit a closed-form solution which restricts the function class prohibitively. 
Alternatively, gradient-based iterative optimization can be employed, which allows for a larger class of functions but 
typically fails to yield satisfactory solutions for complicated function approximators such as deep neural network models.
In addition, calculating first-order or even second-order information for trajectory optimization tends to be computationally costly, which makes it hard to meet the time constraints of real-world settings.
This motivates the usage of zero-order, i.e gradient-free or sample-based  methods, such as the Cross-entropy Method (CEM) that do not rely on gradient information but are efficiently parallelizable.

\begin{figure}[tbh]
\centering
    \begin{subfigure}[b]{0.32\textwidth}
         \centering
         \includegraphics[width=\textwidth]{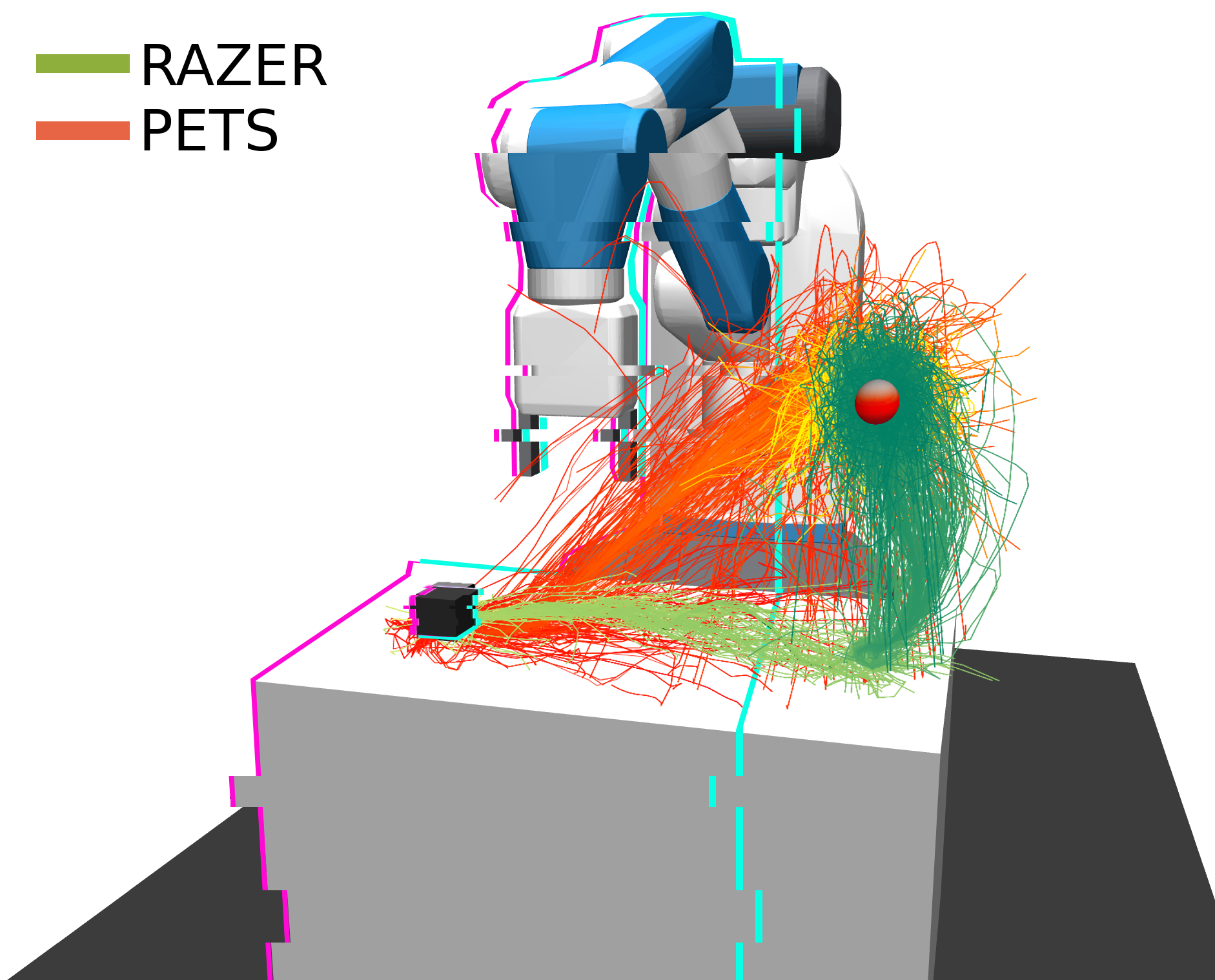}
         \caption{Noisy-FetchPickAndPlace}
         \label{fig:aleatoric-fpp}
    \end{subfigure}
    \hfill
    \begin{subfigure}[b]{0.32\textwidth}
         \centering
         \includegraphics[width=\textwidth]{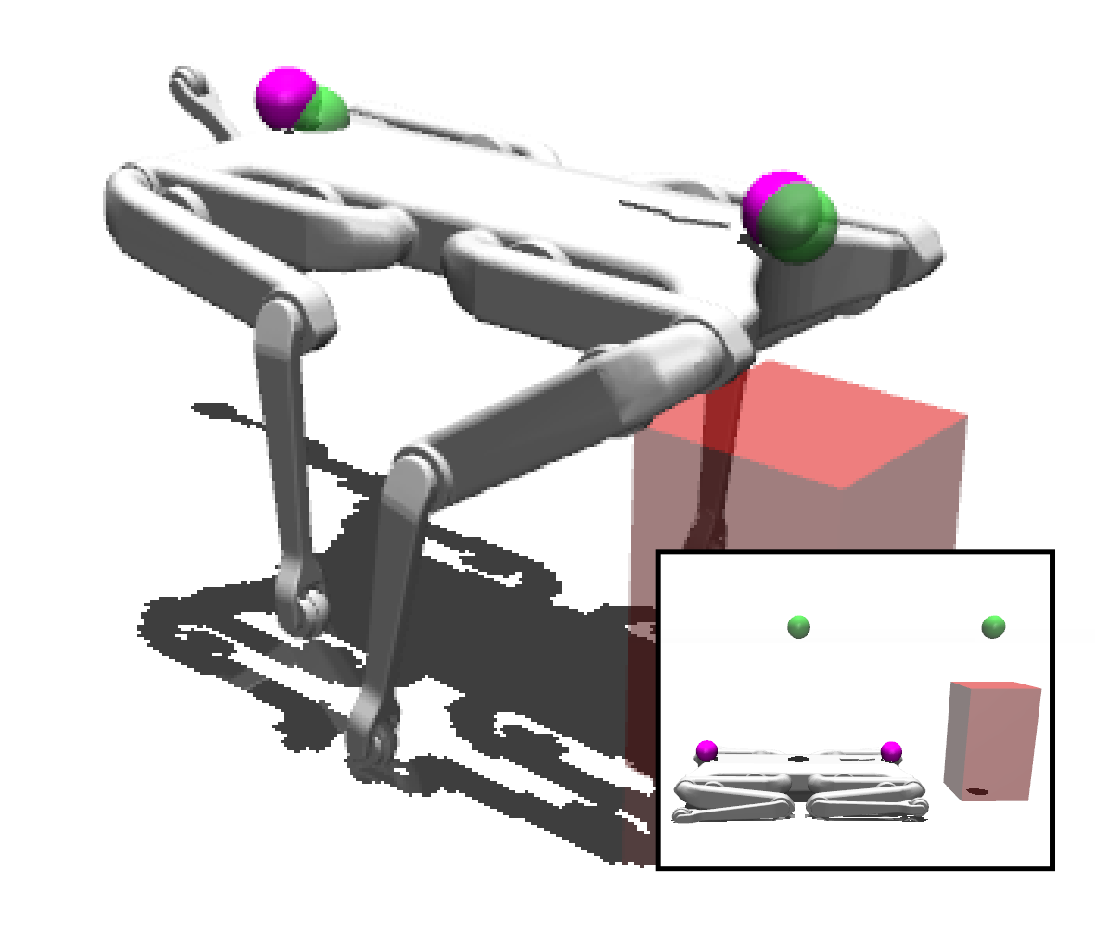}
         \caption{Solo8-LeanOverObject}
         \label{fig:solo8}
    \end{subfigure}\hfill
    \begin{subfigure}[b]{0.3\textwidth}
         \centering
         \includegraphics[width=\textwidth]{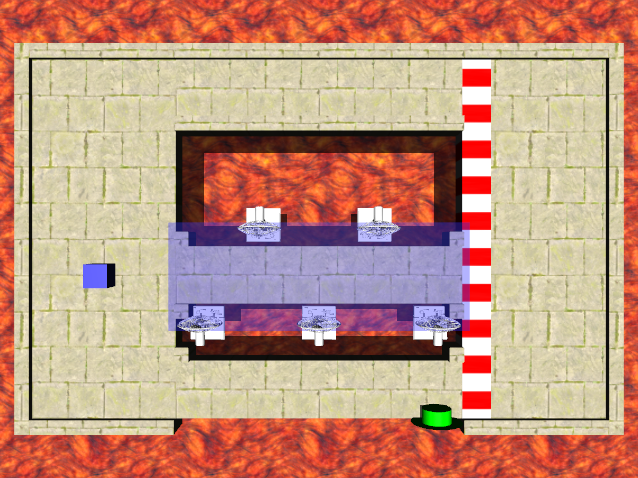}
         \caption{BridgeMaze}
         \label{fig:bridgeMaze}
    \end{subfigure}
    \caption{Environments considered for uncertainty-aware planning. }\label{fig:envs}
\end{figure}

Many methods relying on a learned model and zero-order trajectory optimizers have been proposed~\citep{chua2018:pets,ba2019poplin,Williams2015MPPI}, but all share the same problem: compounding of errors through auto-regressive model prediction.
This naturally brings us to the question of how can we effectively manage model errors and uncertainty to be more data-efficient and safe.
Arguably, this is one of the main obstacles to applying data-driven model-based methods to the real world, \eg to robotics settings.

In this work, we introduce a risk-averse zero-order trajectory optimization method (\method{}) for managing errors and uncertainty in zero-order MPC and test it on challenging scenarios (\fig{fig:envs}).
We argue that it is essential to differentiate between the two types of uncertainty in the model-predictive setting: the aleatoric uncertainty arising from inherent noise in the system and epistemic uncertainty arising from the lack of knowledge~\cite{HORA1996AE, KIUREGHIAN2009AE}.
We measure these uncertainties by making use of probabilistic ensembles with trajectory sampling~\citep{chua2018:pets} (PETS).
Our contributions can be summarized as follows: (i)~method for separation of uncertainties in probabilistic ensembles (termed PETSUS); 
(ii)~efficient use of aleatoric and epistemic uncertainty in model-based zero-order trajectory optimizers; (iii)~an simple but practical approach to probabilistic safety constraints in zero-order MPC.

\section{Related Work} \label{sec:related_work}

\paragraph{Uncertainty Estimation} In the typical model-based reinforcement learning (MBRL) setting, the true transition dynamics function is modeled through an approximator. Impressive results have been achieved by both parametric models~\cite{lenz2015deepmpc, fu2016one, gal2016improving, hafner2019learning}, such as neural networks, and nonparametric models~\cite{kocijan2004gaussian, nguyen2008local, grancharova2008explicit, deisenroth2013gaussian}, such as Gaussian Processes (GP). The latter inspired seminal work on the incorporation of the dynamics model's uncertainty for long-term planning~\cite{deisenroth2013gaussian, kamthe2018data}. However, their usability is limited to low-data, low-dimensional regimes with smooth dynamics~\cite{Rasmussen2003GaussianPI, rasmussen2006gpbook}, which is not ideal for robotics applications. Alternative parametric approaches include ensembling of deep neural networks, used both in the MBRL community~\cite{chua2018:pets, kurutach2018modelensemble}, and outside~\cite{osband2016ensembledqn, lakshminarayanan2017deepensembles}. In particular, ensembles of \emph{probabilistic} neural networks  established state-of-the-art results in the MBRL community~\cite{chua2018:pets}, but focus mainly on estimating the expected cost and disregard the underlying uncertainties.
In comparison, we propose a treatment of the resulting uncertainties of the ensemble model.

\paragraph{Zero-order MPC} 
The learned model can be used for policy search like in PILCO~\cite{deisenroth2011pilco, deisenroth2013gaussian, kamthe2018data, curi20hucrl} or for online model-predictive control (MPC)~\cite{morari1999mpc, williams2017information, chua2018:pets}.
In this work, we do planning in an MPC fashion and employ a zero-order method as a trajectory optimizer, since less sensitive to hyperparameter tuning and less likely to get stuck in local minima of complex objective functions. Specifically, we consider a sample-efficient implementation of the Cross-Entropy method~\cite{Rubinstein99cem,botev2013cem} introduced in~\cite{pinneri2020icem}.
\paragraph{Safe MPC} Separating the sources of uncertainty is of particular importance for AI applications directly affecting humans' safety, as self-driving cars, elderly care systems, or in general any application that involves a physical interaction between the AI system and humans. Disentangling epistemic from aleatoric uncertainty allows for separate optimization of the two, as they represent semantically different objectives: efficient exploration and risk-awareness.
Extensive research on uncertainty decomposition has been done in the Bayesian setting and the context of safe policy search~\cite{mihatsch2002risk, garcia2015comprehensive, depeweg2017learning, depeweg2018decomposition}, MPC planning~\cite{Arruda2017UncertaintyAP, lee2020perceptualmpc, Abraham2020mppiuncertainty}, and distributional RL~\cite{Clements2019EstimatingRA, Zhang2021SafeDR}. 
On the other side, a state-of-the-art baseline for ensemble learning like PETS~\cite{chua2018:pets}, despite estimating both uncertainties, only optimizes for the \emph{expected} cost during action evaluation. Our work aims at filling this gap by explicitly integrating the propagated uncertainty information in the zero-order MPC planner.

\section{Method}\label{sec:method}\vspace{-.8em}

Our approach concerns itself with the efficient usage of uncertainties in zero-order trajectory optimization and is therefore generally applicable to such optimizers.
We are interested in modeling noisy system dynamics $x_{t+1} = f(x_t,u_t,  w(x_t, u_t))$ where $f$ is a nonlinear function, $x_t$ the observation vector, $u_t$ applied control input and $w(x_t, u_t)$ a noise term sampled from an arbitrary distribution.

Consequently, in the absence of prior knowledge about the function $f$, the system needs to be modeled by a complex function approximator such as a neural network.
Furthermore, we are interested in managing uncertainties based on our fitted model, which is erroneous.
To this end, we use stochastic ensembles of size $K$, where the output of each model $\gp^k(x_t, u_t)$ are parameters of a normal distribution depending on input observation $x_t$ and control $u_t$.
As a by-product, our auto-regressive model prediction based on controls $\useq$ becomes a predictive distribution over trajectories $\tau$; $\trajdist(x_t, \useq) \coloneqq p(\tau | x_t, \useq; \theta)$ where $\theta$ denotes the parameters of the ensemble.
For convenience, from this point onward we will differentiate between multiple usages of $\trajdist$.
We denote with $\xtdist$ the distribution $ p(x_{t+\Delta t} | \xinit, \useqinit; \theta)$ over states at time step $t+\Delta t$ and $\gptdist$ the distribution over the Gaussian parameter outputs $p(\gp_{t+\Delta t} | \xinit, \useqinit ; \theta)$ at time step $t+\Delta t$ of the planner. 

\subsection{Planning and Control}

To validate our hypothesis that accounting for uncertainty in the environment and model prediction is essential to develop risk-averse policies, we use the Cross-Entropy Method (CEM) with improvements suggested in~\citet{pinneri2020icem}.
Accordingly, at each time step $t$ we sample a finite number of control sequences $\useq$ for a finite horizon $H$ from an isotropic Gaussian prior distribution which we evaluate from the state $x_t$ using an auto-regressive forward-model and the cost function.
The sampling distribution is refitted in multiple rounds based on good-performing (elite) trajectories.
After this optimization step, the first action of the mean of the fitted Gaussian distribution is executed.
Since this approach utilizes a predictive model for a finite horizon at each time step, it naturally falls into the category of Model Predictive Control (MPC) methods.

Although we use CEM, our approach of managing uncertainty can generically be applied to other zero-order trajectory optimizers such as MPPI~\cite{williams2017information}, by a modification of the trajectory cost function.

\subsection{The Problem of Uncertainty Estimation}

Since we have a stochastic model of the dynamics, at the model prediction time step $t$ we observe a distribution over potential outcomes.
Indeed, since our model outputs are parameters of a Gaussian distribution, with auto-regressive predictions we end up with a distribution over possible Gaussians for a certain time step $t$.

Given a sampled action sequence $\vu$ and the initial state $x_t$ we observe a distribution over trajectories $\psi_\tau$.
To efficiently sample from the trajectory distribution $\psi_\tau$ we use the technique introduced by~\citet{chua2018:pets} (PETS) which involves prediction particles that are sampled from the probabilistic models and randomly mixed between ensemble members at each prediction step. 
In this way, the sampled trajectories are used to perform a Monte Carlo estimate of the expected trajectory cost $\mathop{\E}_{\tau\sim \trajdist}[c(\tau)]$.
However, this does not take the properties of $\trajdist$ into account, which might be a high-entropy distribution and may lead to very risky and unsafe behavior. 
In this work, we alleviate this by looking at the properties of $\trajdist$, \ie different kinds of uncertainties arising from the predictive distribution.

\subsection{Learned Dynamics Model}\label{seq:dyn_model}

We learn a dynamics model ${f_\theta}$ that approximates the true system dynamics $x_{t+1} = f(x_t, u_t, w(x_t, u_t))$. 
As a model class, we use an ensemble of neural networks with stochastic outputs as in~\citet{chua2018:pets}. Each model $k$, parameterizes a multivariate Gaussian distribution with diagonal covariance, $f_\theta^k(x_t,u_t) = \gN(x_{t+1}; x_t+\muk(x_t, u_t), \Sigma_\theta^k(x_t, u_t))$ where $\muk(\cdot, \cdot)$ and $\sigmak(\cdot, \cdot)$ are model functions outputting the respective parameters. 

Iteratively, while interacting with the environment, we collect a dataset of transitions $\mathcal{D}$ and train each model $k$ in the ensemble  by the following negative log-likelihood loss on the Gaussian outputs:
\begin{align}
   \gL(\theta, k) = \E_{x_t,u_t,x_{t+1} \sim \mathcal{D}} &\Big[ 
    - \log \gN(x_{t+1} ; x_t+\muk(x_t, u_t), \Sigma_\theta^k(x_t, u_t)) 
    \Big]
\end{align}
In addition, we use several regularization terms to make the model training more stable.
We provide more details on this in \supp{supp_sec:imp_details}.

\subsection{Separation of Uncertainties} %
\label{sec:uncertainty_estimation}

In the realm of parametric estimators, two uncertainties are of particular interest.
\emph{Aleatoric} uncertainty is the kind that is irreducible and results from inherent noise of the system, \eg sensor noises in robots.
On the other hand, we have \emph{epistemic} uncertainty resulting from lack of data or knowledge which is reducible.
This begs the question, how can we separate these uncertainties given an auto-regressive dynamics model $f_\theta$?
The way that we efficiently sample from $\trajdist$ is by mixing sampled prediction particles, similarly as in PETS\citep{chua2018:pets}.
This process is illustrated by the red lines in Fig.~\ref{fig:ensemble-model}. 

Simple model prediction disagreement is not a good measure for \emph{aleatoric} uncertainty since it can be entangled with epistemic uncertainty.
Given our assumptions about the system dynamics, we measure \emph{aleatoric} uncertainty as the entropy of the predicted normal distributions of the ensemble models. More concretely, given a sampled particle state $\px_t$, we define the estimated aleatoric uncertainty for ensemble model associated to particle $b$ at time step $t$ as:
\begin{equation}\label{eq:aleatoric_uncertainty_particle}
    \ua_b(x | \px_t, u_t) = \gH_{x \sim \xtbdist}(x) 
\end{equation}
Where $\xtbdist$ is the output distribution of ensemble model based on inputs $\px_t$, $u_t$.
Since in the end we are interested in the aleatoric uncertainty incurred from applying the action sequence $\useq$ from initial state $\xinit$, the quantity of interest for us is the expected aleatoric uncertainty for time slice $t$:
\begin{equation}\label{eq:aleatoric_uncertainty_approx}
    \ua(x| u_t) = \E_{ \px_{b} \sim  \xtdist} \Big[ \ua_b(x | \px_{b}, u_t) \Big ]
\end{equation}
Intuitively, because we only have access to the ensemble for sampling,  we take a time-slice in the sampled trajectories from $\trajdist$ and compute the output entropies. 
Moreover, since we assume a Gaussian 1-step predictive distribution this is an expectation over differential Gaussian entropy.
An alternative way of computation which we also explore in this work is calculating the expected particle variance for time slice $t$ of the prediction horizon:
\begin{wrapfigure}[16]{r}{0.59\textwidth}
    \centering
    \includegraphics[width=.98\linewidth]{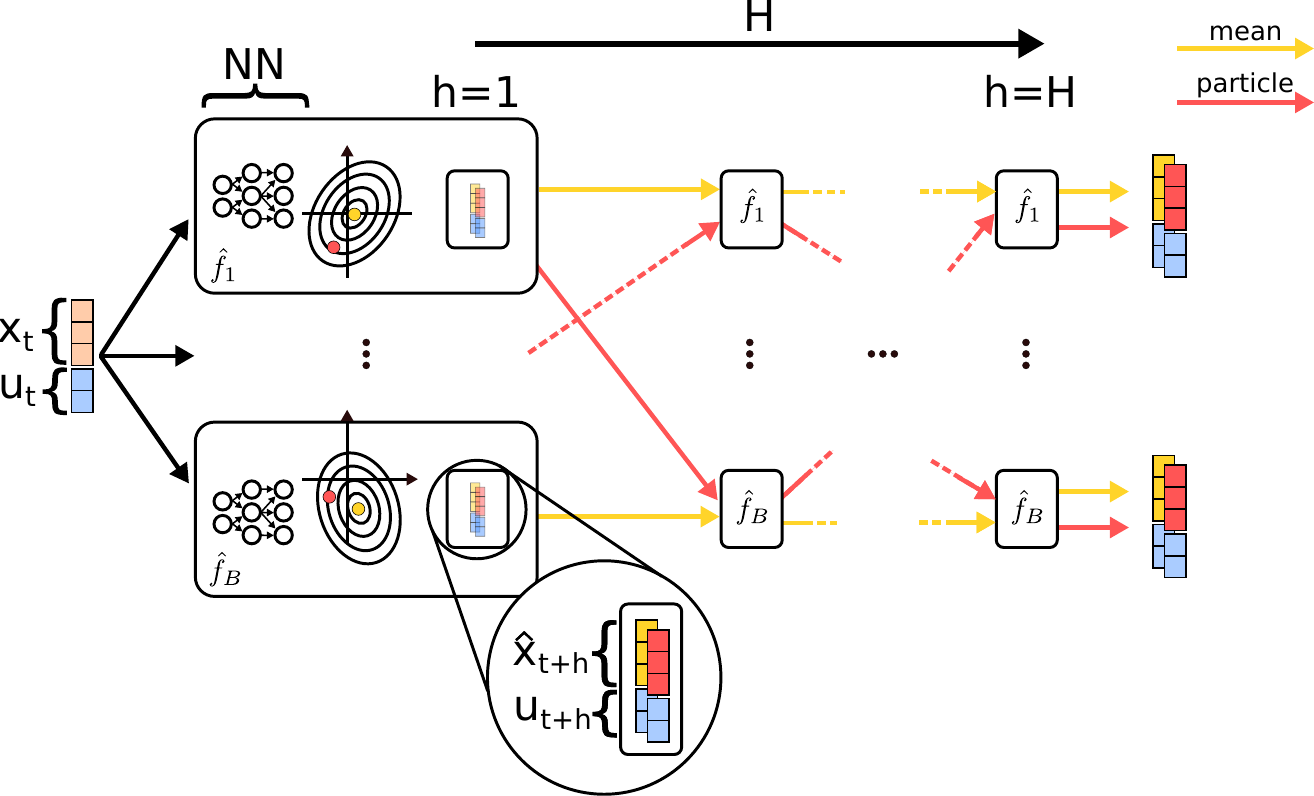}
    \caption{Probabilistic Ensembles with Trajectory Sampling and Uncertainty Separation (PETSUS) }
    \label{fig:ensemble-model}
\end{wrapfigure}
\begin{align} \label{eqn:ensemble:aleatoric}
    \varale_{t+1} = \frac 1 B \sum_{b=1}^B  \sigmak(\px_{t,b},u_t)
\end{align}
For estimating the \emph{epistemic} uncertainty, one would be tempted to look at the disagreement between ensemble models in parameter space $\Var[\theta]$, but this is not completely satisfying, since neural networks tend to be over-parametrized and variance within the ensemble still may exist albeit the optimum has been reached by all ensemble models.
An alternative would be to calculate the Fisher information metric $\gI \coloneqq \Var[\nabla_\theta \log \gL(x_{t+1} | x_t, u_t)]$ where $\gL$ denotes the likelihood function,  but this tends to be expensive to compute.

Given the assumption of local Gaussianity, the true epistemic uncertainty for this case is the predictive entropy over the Gaussian parameters $\gp$ at time step $t+h$.
\begin{equation}
    \ue(\xinit, \useqinit) = \gH_{\gptdist} (\gp \mid \xinit, \useqinit)
\end{equation}
It is easy to verify that this quantity is 0 given perfect predictions of the model.
Note that, because of auto-regressive predictions of a nonlinear model, this is a very difficult object to handle.
Nevertheless, since our predictive distribution $p(x \mid x_t, u_t ; \gp)$ is parametrized by model outputs, we may utilize disagreement in $\gp_t$ to approximate $\ue$. 
To get correct estimations, we need to propagate mean predictions $\mux$ in addition to the particles as illustrated as the yellow lines in Fig. \ref{fig:ensemble-model}.
We  quantify epistemic uncertainty as ensemble disagreement at time step $t$:
\begin{align}
    \varepi(x_{t+1}) = \Var^e[\muk(\mux_t, u_t)] + \Var^e[\sigmak(\mux_t, u_t)] \label{eqn:ensemble:epistemic}
\end{align}
where $\Var^e$ is the empirical variance over the $k=1\ldots K$ ensembles.

\subsection{Probabilistic Safety Constraints}\vspace{-.5em}

When applying data-driven control algorithms to real systems, safety is of utmost importance.
In the realm of zero-order optimization, safety constraints can be easily introduced by putting an infinite cost on constraint-violating trajectories.
Nevertheless, we are dealing with erroneous stochastic nonlinear models which lead to nontrivial predictive distributions of future states, based on the control sequence $\useq$. For this reason, we want to control the risk of violating the safety constraints that we, as practitioners, are willing to tolerate.
If we denote the observation space as $\sX$, given a violation set $\sC \subset \sX$, we define the probability of the control sequence $\vu$ to enter the violation set at time $t+\Delta t$ as $ p(x \in \sC \mid x_t, \useq) =  \int_{x \in \sC} \xtdist(x \mid x_t, \useq)$.
In practice, it is hard to compute this integral efficiently, since our distribution $\xtdist$ is nontrivial as a result of nonlinear propagation of uncertainty.
Furthermore, the violation set $\sC$ might not have the structure necessary to allow an efficient solution to the integral, in which case one needs to resort to Monte Carlo estimation.

To simplify computation and gain speed, we consider box violation sets resulting in each dimension of $x$ being constrained to be outside of $[a, b] \in \{a,b\,|\, a,b \in \sR^2,\, a < b \}$.
By performing moment matching by a Gaussian in each time-slice $\xtdist$, the probability of ending up in state $x$ at time step $t+\Delta t$ is given by integrating $\gN(x ; \mu_{t+\Delta t}, \Sigma_{t+\Delta t})$, where $\mu$ and $\Sigma$ are estimated by Monte Carlo sampling.
If we further assume a diagonal covariance $\Sigma$, this integral can be deconstructed into $d$ univariate Gaussian integrals, which can be computed fast and in closed form (error function).
Hence, the probability of a constraint violation happening at time step $t$ is defined by:
\begin{equation}
    p(x\in \sC \mid x_t, \useq) = \prod_{i=0}^d\int_{x \in \sC} \gN(x^i; \mu^i_{t+\Delta t}, \sigma^i_{t+\Delta t})\label{eqn:safety}
\end{equation}

\subsection{Implementing Risk-Averse ZERo-Order Trajectory Optimization (\method)}\label{sec:algorithm}\vspace{-.5em}
We assume the task definition is provided by the cost $c(x_t,\vu)$.
For trajectory optimization, we start from a state $x_t$ and predict with an action sequence $\useq$
 the future development of the trajectory $\tau$. %
Along this trajectory, we want to compute a single cost term  which is conveniently defined as the expected cost of all particles $\px$ summed over the planning horizon $H$:
\begin{equation}
    c(x_t,\useq) = \sum_{\Delta t=1}^H \frac 1 B \sum_{b=1}^B c(\px^{b}_{t+\Delta t}, u_{t+\Delta t}). \label{eqn:maincost}
\end{equation}
The optimizer, in our case CEM, will optimize the action sequence $\useq$ to minimize the cost in a probabilistic sense, \ie $p(\useq\mid x) \propto \exp(- \beta\, c(x,\useq))$ where $\beta$ reflects the strength of the optimizer (the higher the more likely it finds the global optimum). 
To make the planner uncertainty-aware, we need to make sure it avoids unpredictable parts of the state space by making them less likely. 
Using the aleatoric uncertainty provided by PETSUS \eqn{eqn:ensemble:aleatoric}, we define the aleatoric penalty as 
\begin{equation}
    c_{\ua}(x_t, \useq) =  w_{\ua} \cdot \sum_{\Delta t=1}^{H} \sqrt{\varale_{t+\Delta t}},
    \label{eq:aleatoric_cost}
\end{equation}
where $w_\ua>0$ is a weighting constant. The larger the aleatoric uncertainty, the higher the cost.

To guide the exploration to states where the model has epistemic uncertainty  \eqn{eqn:ensemble:epistemic} (due to lack of data), we use an epistemic bonus:
\begin{equation}
    c_{\ue}(x_t,\useq) = - w_{\ue} \cdot \sum_{\Delta t=1}^{H} \sqrt{\varepi_{t+\Delta t}},
    \label{eq:epistemic_cost}
\end{equation}
where $w_{\ue}>0$ is a weighting constant.
To be able to operate on a real system, the most important part is to adhere to safety constraints. 
As formulated in \eqn{eqn:safety}, the predicted safety violations need to be uncertainty aware, independent of the source of uncertainty.
We integrate this into the planning method by adding:
\begin{equation}
    c_{\sav}(x_t,\useq) = w_{\sav} \cdot \sum_{\Delta t=1}^{H} \big\llbracket p(\hat x_{t+\Delta t} \in \sC) > \delta \big\rrbracket
    \label{eq:safety_cost}
\end{equation}
where $\llbracket\cdot \rrbracket$ is Iverson bracket %
 and $w_{\sav}$ is either a large penalty $c_\mathrm{max}$ or 0 to disable safety. 
An alternative for implementing safety constraints into CEM is by changing the ranking function~\cite{WenUfuk2018:CCM}.
The overall algorithm used in a model-predictive control fashion is outlined in \supp{supp_sec:algo}.

\section{Experiments}\label{sec:experiments}\label{sec:results}

We study our uncertainty-aware planner in $4$ continuous state and action space environments and compare to naively optimizing the particle-based estimate of the expected cost similarly to~\citet{chua2018:pets}.
We start by giving a description of the environments.

\textbf{BridgeMaze} This toy environment (see \fig{fig:bridgeMaze}) was specifically designed to study the different aspects of uncertainty independently. The agent (blue cube) starts on the left platform and has to reach the goal platform on the right. 
To reach the goal platform, the agent has to move over one of three bridges without falling into the lava. 
The upper bridge is safeguarded by walls; hence, it is the safest path to the goal but also the longest. 
The lower bridge has no walls and therefore is more dangerous for an unskilled agent to cross but the path is shorter. 
The middle bridge is the shortest path to the goal. However, randomly appearing strong winds perpendicular to the bridge might cause the agent to fall off the bridge with some probability, making this bridge dangerous.

\textbf{Noisy-HalfCheetah} This environment is based on \textit{HalfCheetah-v3} from the OpenAI Gym toolkit. We introduce aleatoric uncertainty to the system by adding Gaussian noise $\xi \sim \mathcal{N}(\mu, \sigma^{2})$ to the actions when the forward velocity is above $6$. The action noise translates into a non-Gaussian and potentially very complicated state space noise distribution that makes the control problem very challenging.  

\textbf{Noisy-FetchPickAndPlace} 
Based on the \textit{FetchPickAndPlace-v1} gym environment. Additive action noise is applied to the gripper so that its grip on the box might become tighter or looser. The noise is applied for $x$-positions $<0.8$ which is illustrated in \fig{fig:aleatoric-fpp} by a blue line causing the agent to drop the box with high probability if it tries to lift the box too early. 

\textbf{Solo8-LeanOverObject} In this robotic environment, the task of a quadrupedal robot~\citep{grimminger2020open} is to stand up and lean forward to reach a target position (purple markers need to reach green dots in \fig{fig:solo8}) without hitting an object visualized by the red cube representing the unsafe zone. The robot starts in a laying position as shown in the inset of \fig{fig:solo8}. As in the \textit{Noisy-HalfCheetah} environment, Gaussian action noise is applied to mimic real-world perturbances. 

\subsection{Algorithmic Choices and Training Details}\label{sec:algorithmic_choices}

For model-predictive planning we use the CEM implementation from~\citet{pinneri2020icem}. Further details about hyperparameters can be found in \supp{supp_sec:controller}.  For planning, we use the same architecture for the ensemble of probabilistic models, both in \method{} and in PETS. The only difference is that in \method{} we also forward propagate the mean state predictions in addition to the sampled state predictions. Further details can be found in \supp{supp_sec:model_learning}.

For training the predictive model, we alternate between two phases: data collection and model fitting. In the \textit{BridgeMaze} environment, we collect $5$ rollouts of length $80$ steps and append them to the previous rollouts. Afterwards, we fit the model for $25$ epochs. For \textit{Noisy-HalfCheetah}, we collect $1$ rollout and fit for $50$ epochs. For Noisy-FetchPickAndPlace and Solo8-LeanOverObject we replace the $\hat{f}$ in \fig{fig:ensemble-model} with independent instances of noisy ground truth simulators.

Next, we will present \method{}'s exploration and safety behavior in the \textit{BridgeMaze} environment. Afterwards, we are going to discuss planning with external safety constraints in the \textit{Solo8-LeanOverObject} environment. We complete this section with results on \textit{Noisy-HalfCheetah} and \textit{Noisy-FetchPickAndPlace}. 

\subsection{Active Learning for Model Improvement}\label{sec:active_learning}

\begin{figure}
    \centering
    \captionsetup[subfigure]{position=b}
    \begin{subfigure}[t]{.32\textwidth}
            \includegraphics[width=\textwidth]{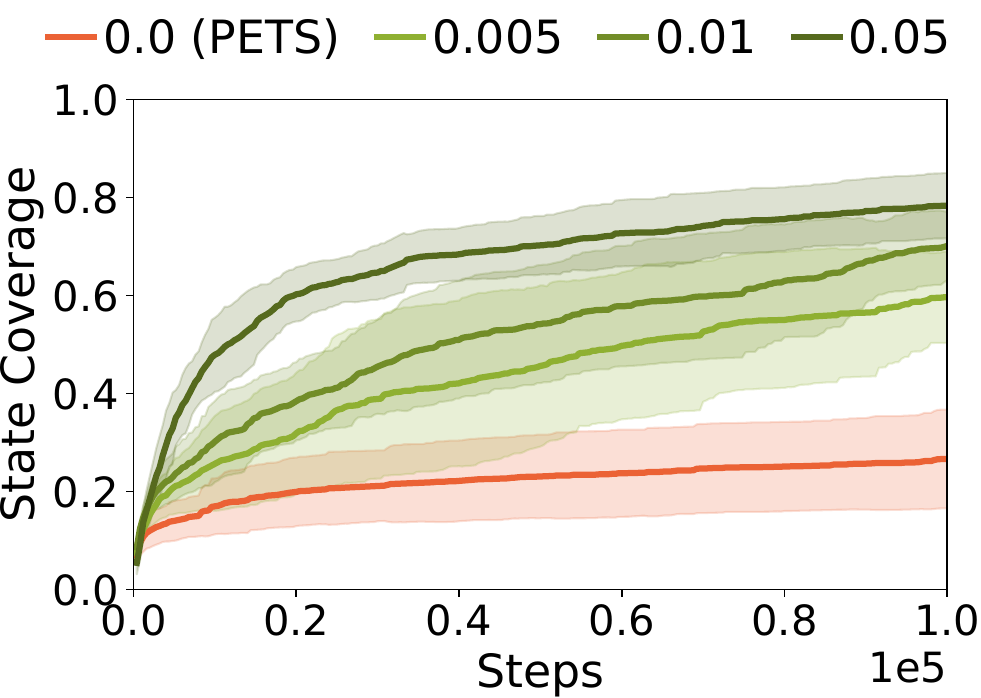}\vspace{-.07em}
        \subcaption{State space exploration over time depending on epistemic bonus ($w_\ue$).}
        \label{fig:exploration_over_time}
    \end{subfigure}
    \hfill
    \begin{subfigure}[t]{.28\textwidth}
       \includegraphics[width=\textwidth]{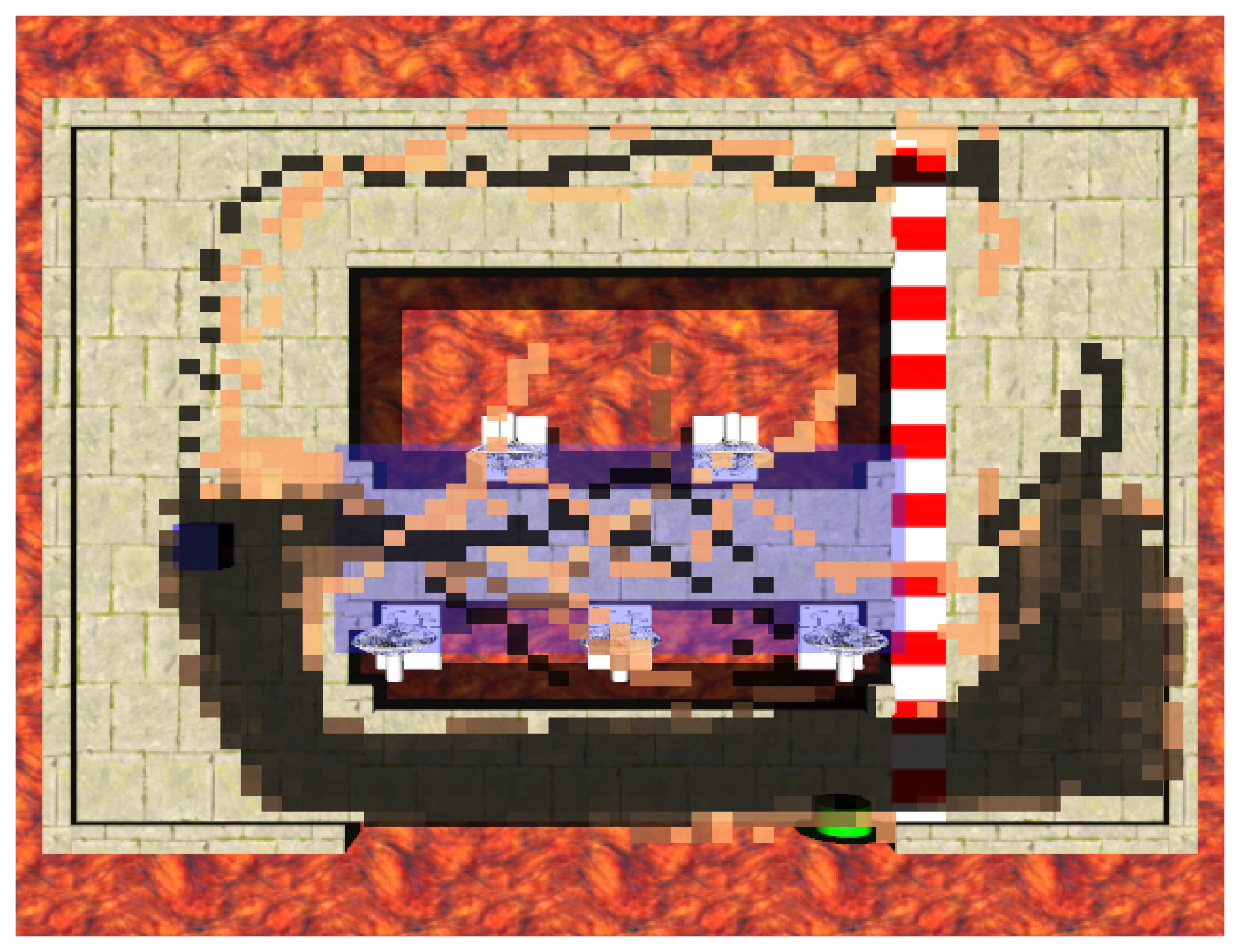}
        \subcaption{State space coverage with $w_\ue=0$.}
        \label{fig:state_coverage_pets}
    \end{subfigure}
    \hfill
    \begin{subfigure}[t]{.28\textwidth}
       \includegraphics[width=\textwidth]{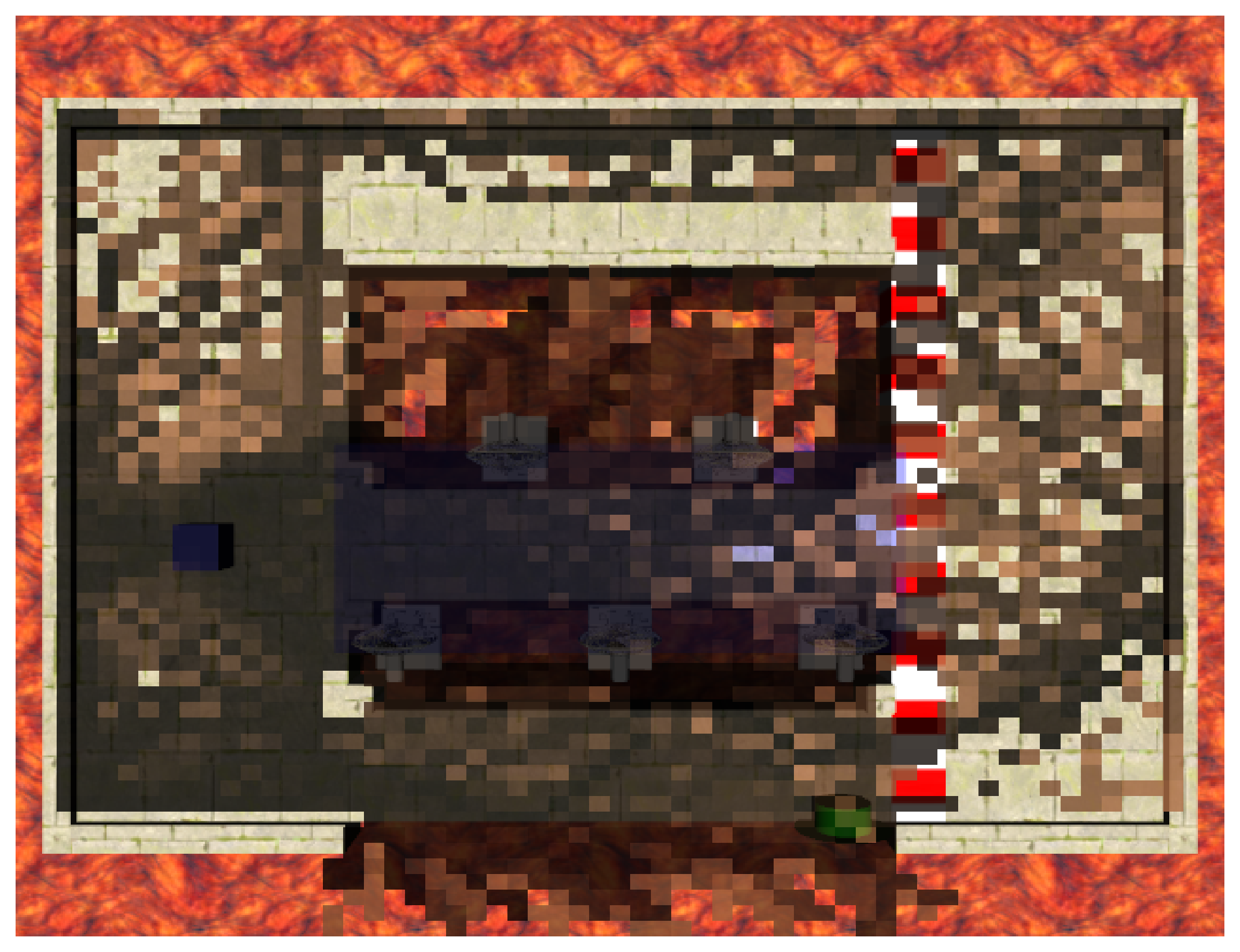}
        \subcaption{State space coverage  with $w_{\ue}=0.05$.}
        \label{fig:state_coverage_ours}
    \end{subfigure}\vspace{-.5em}
    \caption{Active learning setting: The epistemic bonus allows \method{} to seek states for which no or only little training data exists (a,c). Means and standard deviations for (a) were computed over 5 runs. PETS overfits to a particular solution (b). 
    In (b) and (c), the brightness of the dots is proportional to the time when they were first encountered.}
    \label{fig:exploration}
\end{figure}

If model uncertainties are used for risk-averse planning, they are only meaningful if the model has the right training data. Only from good data can the parameters of the approximate noise model be learned correctly. In case of too little data, the agent might avoid parts of the state space due to an overestimation of the model uncertainties. On the other hand, the agent might enter unsafe regions for which the uncertainties are underestimated. By adding the epistemic bonus to our domain-specific cost, the planner can actively seek states with high epistemic uncertainty, \ie for which no or only little training data exists.

\Fig{fig:pointBridgeAleatoric} shows this active data gathering process for the \textit{BridgeMaze} environment. PETS finds one particular solution to the problem of reaching the goal platform. It chooses the path over the safer, lower bridge rather than the dangerous middle path and the longer path via the upper bridge (\fig{fig:state_coverage_pets}). Once, one solution is found, the model overfits to it without exploring any other parts of the state space. This is also reflected in the plateauing of the red curve in \fig{fig:exploration_over_time}. 

In comparison, \method{} actively explores larger and larger parts of the state space with an increasing weight of the epistemic bonus (\fig{fig:exploration_over_time}). \method{} not only finds the easy solution found by PETS but also extensively explores other parts of the state space (\fig{fig:state_coverage_ours}). To not get stuck at the middle bridge during exploration due to the inherent noise, it is important to separate between epistemic and aleatoric uncertainties. Only the former should be used for exploration. With enough data, our model can correctly capture the uncertainties of these states resulting in the epistemic uncertainty approaching zero.

\subsection{Risk-Averse Planning}\label{sec:risk_averse_planning}

\begin{figure}
    \centering
    \captionsetup[subfigure]{position=b}
    \begin{subfigure}[t]{.48\textwidth}\centering
             \includegraphics[width=.75\textwidth]{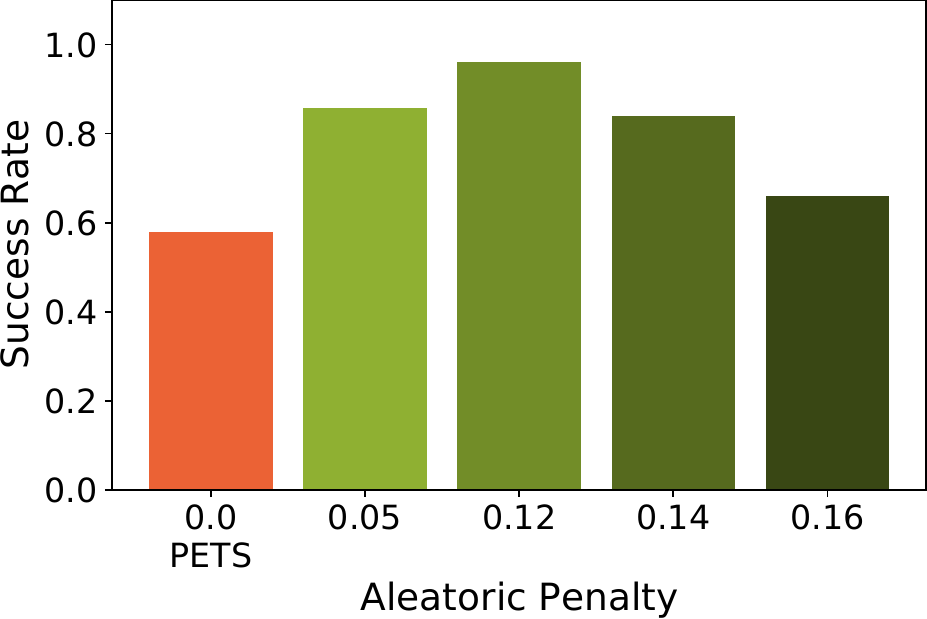}
            \subcaption{\textit{BridgeMaze} success depending on $w_{\ua}$ for $50$ runs.}
            \label{fig:pointBridgeAleatoric}
    \end{subfigure}    \hfill
    \begin{subfigure}[t]{.48\textwidth}\centering
        \includegraphics[width=.8\linewidth]{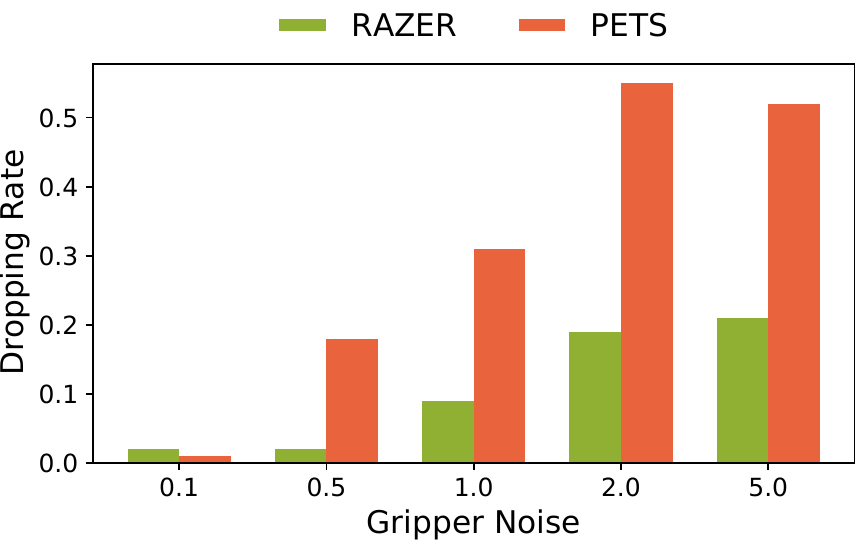}
        \subcaption{Dropping rate in \textit{Noisy-FetchPickAndPlace} for $100$ runs.}
        \label{fig:fpp_droppingrate}
    \end{subfigure}\vspace{-.5em}
    \caption{Risk-averse planning in the face of aleatoric uncertainty yields higher success rates in noisy environments. For (b) we use ground truth models  and a fixed aleatoric penalty weight $w_{\ua}$.
    }
    \label{fig:my_label}
\end{figure}

Once a good model is learned, it can be used for safe planning. What differentiates \method{} from PETS is that it makes explicit use of uncertainty estimates while in the latter uncertainties only enter planning by taking the mean over the particle costs and not differentiating between different sources of uncertainty. 

\paragraph{BridgeMaze} \Fig{fig:pointBridgeAleatoric} shows the success rate of PETS and \method{} in the \textit{BridgeMaze}. In both cases, we use the same model that was trained from data collected during a training run with $w_{\ue}=0.05$. Hence, the model saw enough training data from all parts of the state space. The noise in the environment is tuned such that there is a chance to cross the bridge without falling. While in \fig{fig:state_coverage_pets} PETS avoided this path because of an overestimation of the state's value due to a lack of training data and sometimes sees a chance to cross the bridge. However, these attempts are very likely to fail because of stronger winds that occur randomly, resulting in a success rate of only $58\%$. \method{} does not rely on sampling for the aleatoric part and can thus avoid risk. With a higher penalty constant the success rate increases up to $96\%$ but only as long as the agent is willing to take a risk at all. For large values of $w_{\ua}$ %
the agent becomes so conservative that it only moves slowly (decreasing reward in \fig{fig:pointBridgeAleatoric}). %

\begin{figure}
    \centering
    \includegraphics[width=0.6\linewidth]{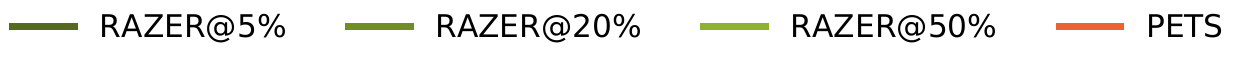}\\
    \captionsetup[subfigure]{position=b}
    \begin{subfigure}{.36\textwidth}
             \includegraphics[height=3cm]{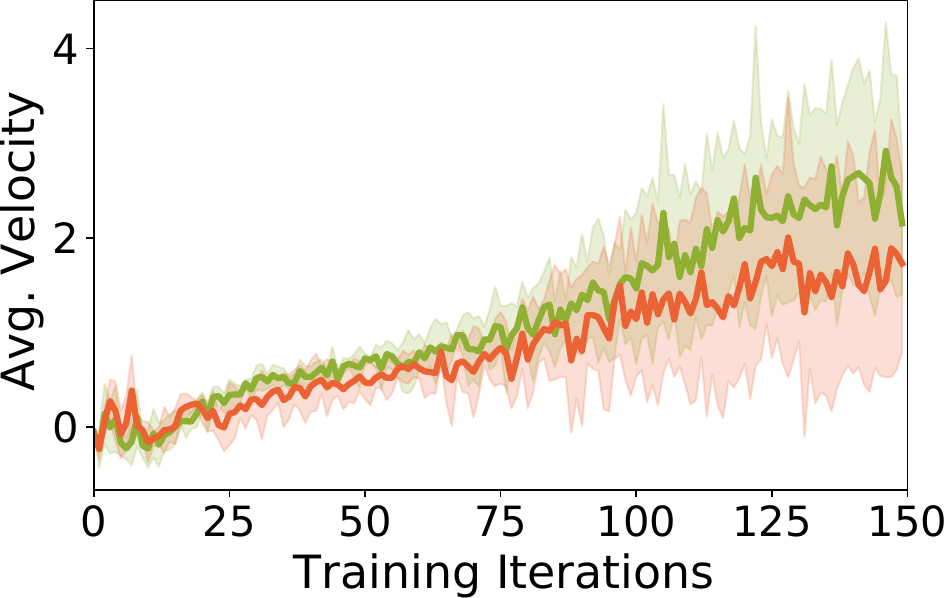}
            \subcaption{With aleatoric penalty (10 runs).}\label{fig:halfcheetah-aleatoric}
    \end{subfigure}
    \begin{subfigure}{.36\textwidth}
            \centering
           \includegraphics[height=3cm]{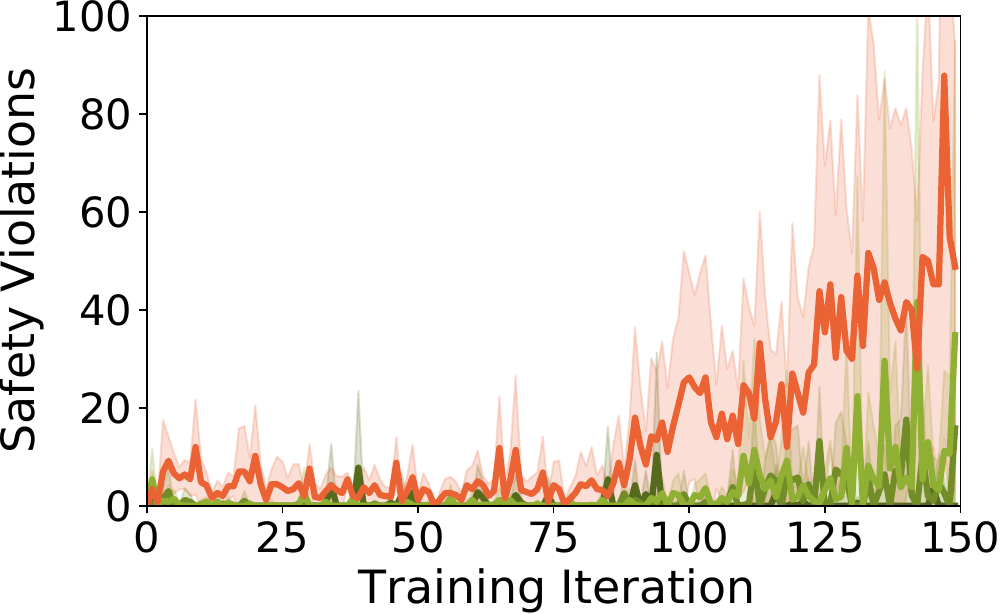}
            \subcaption{With safety constraints.}\label{fig:halfcheetah-safety}
    \end{subfigure}
     \begin{subfigure}{.24\textwidth}
            \centering
            \raisebox{0.5em}{
                \includegraphics[width=\textwidth]{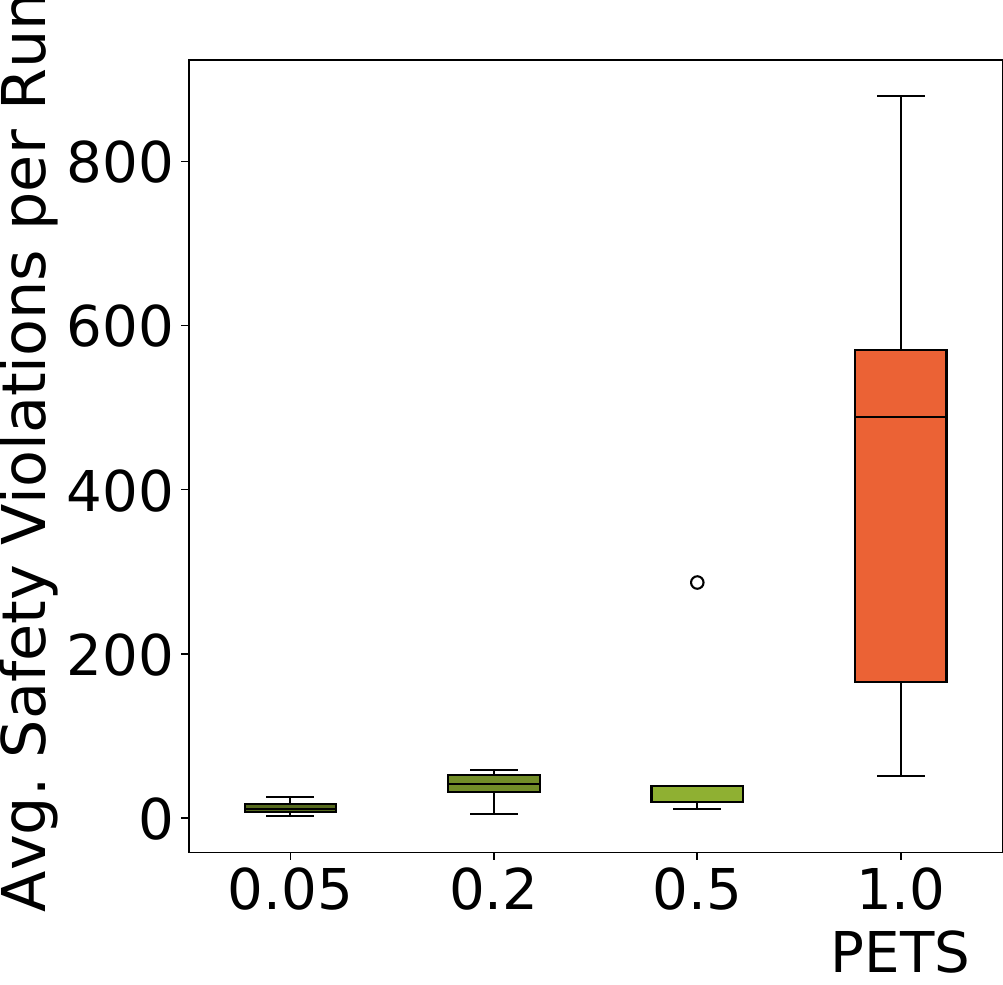}
            }\vspace{-1em}
            \subcaption{With safety constraints.}\label{fig:halfcheetah-safety-box}
    \end{subfigure}\vspace{-.5em}
    \caption{\textit{Noisy-HalfCheetah} environment (task lengths 300 steps) with learning models from scratch. 
    At 150 iterations we have seen only 45k points. 
    (a) Performance under noisy actions. By applying the aleatoric penalty, \method{} can navigate the uncertainties better -- leading to higher returns faster. (b) Safety violations above a certain body height (simulating a low ceiling) for different values of $\delta$. 
    With increasing $\delta$, \method{} is seldomly violating constraints in stark contrast to PETS. In (c) the number of violations is averaged over the last 50 iterations (summed over 10 rollouts).}
    
\end{figure}

\paragraph{Noisy-HalfCheetah}
How does \method{} perform on the \textit{Noisy-HalfCheetah} environment when models are learned from scratch?
Without aleatoric penalty, the planner is optimistic. Risky situations are only detected if a failing particle is sampled. 
Thus, the noise is mostly neglected and the robot increases its velocity, gets destabilized, and ends up slower than with the aleatoric penalty (\fig{fig:halfcheetah-aleatoric}).

\paragraph{Noisy-FetchPickAndPlace}
In this environment, a 7-DoF robot arm should bring the box to a target position --
starting and target positions are at the opposite sides of the table. 
The shortest path is to lift the box and move in a straight line to the target. 
However, with noise applied to the gripper action, there is a certain probability to drop the box along the way. 
When penalizing aleatoric uncertainty, this is avoided and also fewer trajectory samples are ``wasted'' in high-entropic regions, as presented in \fig{fig:aleatoric-fpp}. 
\Fig{fig:fpp_droppingrate} shows the number of times the box is dropped on the table depending on the aleatoric penalty. 
\method{} adopts a cautious behavior, preferring to slide the box on the table and lifting it only in the area without action noise, achieving a dropping rate lower than 20\%, even when considerable noise is applied.

\subsection{Planning with External Safety Constraints}\label{sec:safty_constraint}

\paragraph{Noisy-HalfCheetah:} We consider a safety constraint on the height of the body above ground simulating a narrow passage.
\Fig{fig:halfcheetah-safety} shows the number of safety violations. Note that PETS has the same penalty cost for hard violations.

\begin{figure}
    \centering
    \begin{tabular}{c@{\hspace{4em}}c}        
             \includegraphics[width=.35\linewidth]{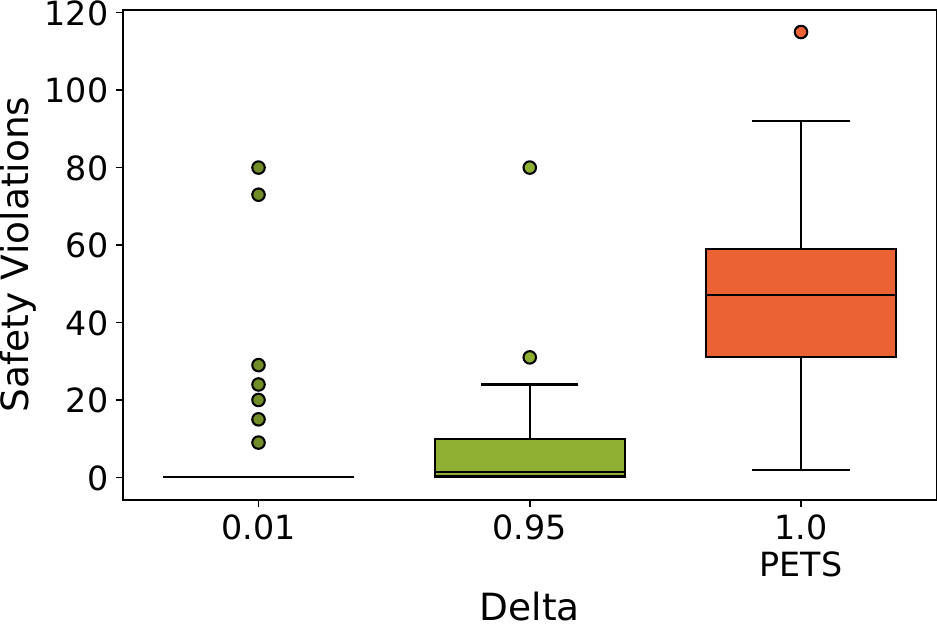}&     \includegraphics[width=.35\linewidth]{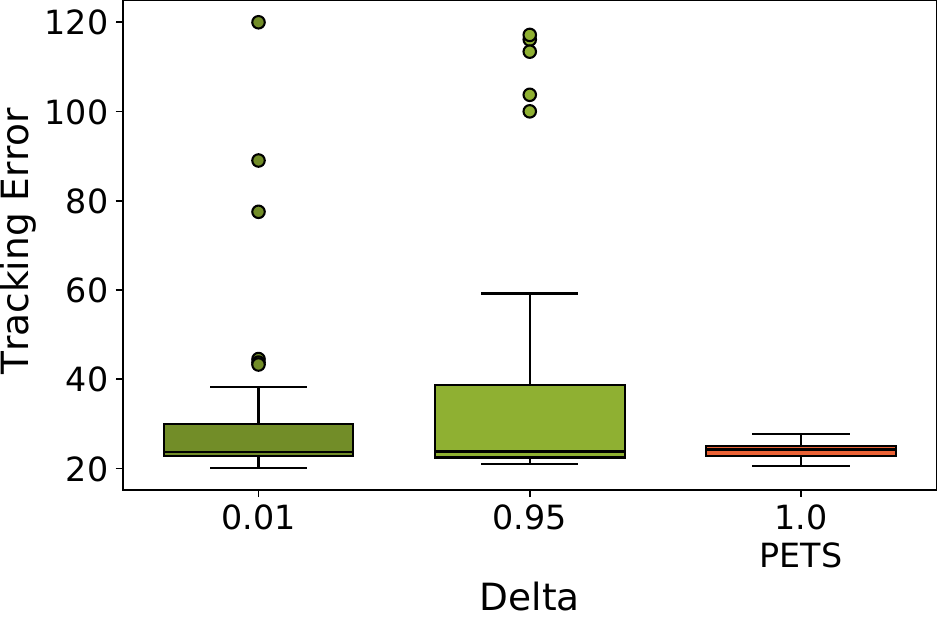}
    \end{tabular}\vspace{-.5em}
    \caption{Safe planning vs.\ task-oriented planning in the \textit{Solo8-LeanOverObject} environment with noisy actions. Left: number of safety violations for different values of $\delta$ (\eqn{eq:safety_cost}). Right: enforcing safety constraints causes slight reduction in tracking accuracy due to the fixed planning budget and the competing objectives of task and safety costs. }
    \label{fig:solo:performance}
\end{figure}

\paragraph{Solo8-LeanOverObject:} In this experiment, the robot has to move to two target points with its front and rear of the trunk while avoiding entering a specified rectangular area (fragile object). The front feet are fixed. To track the points, the robot has to lean forward, such that it can lose balance due to noisy actions. 
In contrast to PETS, \method{} successfully manages to satisfy the safety constraints almost always as shown in \fig{fig:solo:performance}.
 However, satisfying the safety constraint comes with the cost of reduced tracking accuracy. %

\section{Conclusion} \label{sec:discussion}\vspace{-1em}

In this work, we have provided a methodology to separate uncertainties in stochastic ensemble models (PETSUS)  which can be used as a tool to build risk-averse model-based planners that are also data-efficient and enforce safety through probabilistic safety constraints (\method).
This type of risk-averseness can be achieved by a simple modification of the cost function in form of uncertainty penalties in zero-order trajectory optimizers.

Furthermore, the separation of uncertainties allows us to do proper exploration via epistemic bonus which benefits generalization of the model and therefore makes it applicable to more settings. As future work, it would be of interest to see this approach applied to a proper transfer learning setting from simulations to real systems, where risk-averseness combined with exploratory behavior is crucial for efficient learning and safe operation.

\section{Acknowledgments}

The authors thank the International Max Planck Research School for Intelligent Systems (IMPRS-IS) for supporting Marin Vlastelica and Sebastian Blaes.
We acknowledge the support from the German Federal Ministry of Education and Research (BMBF) through the Tübingen AI Center (FKZ: 01IS18039B).
Georg Martius is a member of the Machine Learning Cluster of Excellence, EXC number 2064/1 – Project number 390727645.

\bibliography{main}

\appendix

In this supplementary we provide additional details for our method.
We also provide videos that showcase risk-averse behavior of RAZER at \url{https://sites.google.com/view/razer-traj-opt}.

Our research suffered from pandemic impacts on lab access which is detailed in \sec{sec:suppl:pandemic}.

\section{Implementation Details}\label{supp_sec:imp_details}

\subsection{Model Learning}\label{supp_sec:model_learning}

Parameters used for model learning in the \textit{BridgeMaze} experiments.

\begin{table}[h]
    \caption{Model parameters}
    \centering
    \begin{minipage}[t]{0.49\linewidth}
    \centering
    \small
    Ensemble parameters\\[.5em]
        \begin{tabular}{l|l}
            \toprule
            Name & Value\\
            \midrule
            num\_layers & $6$ \\
            size & $400$\\
            activation & silu\\
            ensemble\_size (n) & $5$\\
            output\_activation & None\\
            l1\_reg & $0$\\
            weight\_initializer & truncated\_normal\\
            bias\_initializer & $0$\\
            use\_spectral\_normalization & False\\
            \bottomrule
        \end{tabular}\\[1em]
    
    Stochastic NN parameters\\[.5em]
    \begin{tabular}{l|l}
            \toprule
            Name & Value\\
            \midrule
            var\_clipping\_low & $-10.0$ \\
            var\_clipping\_high & $4$ \\
            state\_dependent\_var & True\\
            regularize\_automatic\_var\_scaling & False\\
             \bottomrule
        \end{tabular}
    \end{minipage}
    \hfill
    \begin{minipage}[t]{0.49\linewidth}
    \centering
    \small
    Remaining parameters\\[.5em]
        \begin{tabular}{l|l}
            \toprule
            Name & Value\\
            \midrule
            lr & $0.002$ \\
            grad\_norm & $2.0$ \\
            batch\_size & $512$ \\
            weight\_decay & $1e^{-5}$ \\
            use\_input\_normalization & True \\
            use\_output\_normalization & False \\
            epochs & $25$ \\
            predict\_deltas & True \\
            train\_epochs\_only\_with\_latest\_data & False \\
            iterations & $0$ \\
            optimizer & Adam \\
            propagation\_method & TS1 \\
            sampling\_method & sample \\
             \bottomrule
        \end{tabular}
    \end{minipage}
    \label{tab:model_params}
\end{table}

We bound the predicted log variance by applying (as in \cite[A.1]{chua2018:pets})

\begin{verbatim}
    logvar = max_logvar - softplus(max_logvar - logvar)
    
    logvar = min_logvar + softplus(logvar - min_logvar)
\end{verbatim}

to the output of the network that predicts the log variance, \texttt{logvar}.
In principle, we could differentiate through this bound to automatically adjust the bounds \texttt{max\_logvar} and \texttt{min\_logvar}. However, we decided to not make these parameters learnable.

Parameters used for model learning in the \textit{Noisy-HalfCheetah} environment (only differences to \textit{BridgeMaze} environment).

\begin{table}[h]
    \caption{Model parameters}
    \centering
    \begin{minipage}[t]{.49\linewidth}
    \centering
    \small
    \small
    Ensemble parameters\\[.5em]
    \begin{tabular}{l|l}
        \toprule
        Name & Value\\
        \midrule
        num\_layers & $4$ \\
        size & $200$\\
        \bottomrule
    \end{tabular}\\[1em]
    \small
    Stochastic NN parameters\\[.5em]
    \begin{tabular}{l|l}
        \toprule
        Name & Value\\
        \midrule
        var\_clipping\_low & $-6.0$ \\
        state\_dependent\_var & True\\
         \bottomrule
    \end{tabular}
    \end{minipage}
    \begin{minipage}[t]{.49\linewidth}
    \centering
    \small
    Remaining parameters\\[.5em]
    \begin{tabular}{l|l}
        \toprule
        Name & Value\\
        \midrule
        lr & $0.0002$ \\
        grad\_norm & None \\
        batch\_size & $256$ \\
        weight\_decay & $3e^{-5}$ \\
        epochs & $50$ \\
         \bottomrule
    \end{tabular}
    \end{minipage}
    \label{tab:params:HC}
\end{table}

\subsection{Controller Parameters}\label{supp_sec:controller}

Parameters used in the CEM controller. For an explanation of the different parameters, we refer the reader to\citep{pinneri2020icem}.

\begin{table}[h]
    \caption{Controller parameters, BridgeMaze environment.}
    \centering
    \begin{minipage}[t]{0.49\linewidth}
    \centering
    \small
    Action sampler parameters\\[.5em]
        \begin{tabular}{l|l}
            \toprule
            Name & Value\\
            \midrule
            alpha & $0.1$ \\
            colored\_noise & true\\
            elite\_size & $10$\\
            execute\_best\_elite & true\\
            finetune\_first\_action & false\\
            fraction\_elites\_reused & $0.3$\\
            init\_std & $0.5$\\
            keep\_previous\_elites & true\\
            noise\_beta & $2.0$\\
            opt\_iterations & $3$\\
            relative\_init & true\\
            shift\_elites\_over\_time & true\\
            use\_mean\_actions & true\\
            \bottomrule
        \end{tabular}
    \end{minipage}
    \hfill
    \begin{minipage}[t]{0.49\linewidth}
    \centering
    \small
    Remaining parameters\\[.5em]
        \begin{tabular}{l|l}
            \toprule
            Name & Value\\
            \midrule
            cost\_along\_trajectory & sum \\
            delta & $0.0$ \\
            factor\_decrease\_num & $1$ \\
            horizon & $30$ \\
            num\_simulated\_trajectories & $128$ \\
             \bottomrule
        \end{tabular}
    \end{minipage}
    \label{tab:controller_params}
\end{table}

\begin{table}[h]
    \caption{Controller parameters, Noisy-HalfCheetah environment (only difference to BridgeMaze environment).}
    \centering
    \begin{minipage}[t]{0.49\linewidth}
    \centering
    \small
    Action sampler parameters\\[.5em]
        \begin{tabular}{l|l}
            \toprule
            Name & Value\\
            \midrule
            noise\_beta & $0.25$\\
            opt\_iterations & $4$\\
            \bottomrule
        \end{tabular}
    \end{minipage}
    \hfill
    \begin{minipage}[t]{0.49\linewidth}
    \centering
    \small
    Remaining parameters\\[.5em]
        \begin{tabular}{l|l}
            \toprule
            Name & Value\\
            \midrule
            num\_simulated\_trajectories & $120$ \\
             \bottomrule
        \end{tabular}
    \end{minipage}
    \label{tab:controller_params_hc}
\end{table}

\begin{table}[h]
    \caption{Controller parameters, Solo8-LeanOverObject environment (only difference to BridgeMaze environment).}
    \centering
    \small
    Action sampler parameters\\[.5em]
        \begin{tabular}{l|l}
            \toprule
            Name & Value\\
            \midrule
            init\_std & $0.3$\\
            noise\_beta & $3.0$\\
            \bottomrule
        \end{tabular}
\end{table}

\subsection{Timings}\label{supp_sec:timings}

While our code is not tuned for speed specifically, in this section we provide some timings for a single step in the environment (hyper-parameters are set as specified in \supp{supp_sec:model_learning} and \supp{supp_sec:controller}, with num\_simulated\_trajectories $= 128$ and op\_iterations $= 3$) in \tab{tab:timing}.  

\begin{table}
    \caption{Timings per one environment step in ms.  We measured the timings on a system with $1$ GeForce GTX 1050 Ti, an Intel Core i7-6800K and 31GB of memory.
\label{tab:timing}}
    \centering
    \small
        \begin{tabular}{c|c}
            \toprule
            Environment & Timing [ms]\\
            \midrule
            BridgeMaze & 0.25\\
            Noisy-HalfCheetah & 0.14\\
            \bottomrule
        \end{tabular}
\end{table}

\subsection{Uncertainty Separation}\label{supp_sec:UncertaintySeparation}

In our method, we separate the epistemic uncertainty, denoted as $\ue$ and aleatoric uncertainty, denoted as $\ua$, the details of which are explained in Sec. \ref{sec:method} with the resulting costs that arise.
Since we are using a variant of the CEM algorithm that needs to sort the sampled action sequences $\vu$ according to their cost, the cost of an action sequence is a single floating point number.

The stochastic NN ensemble that we are using samples trajectories from the predictive distribution $\psi_\tau$ for each action sequence $\vu$. 
In addition, our variant (PETSUS), also propagates the mean prediction $\bar x_t$ for each ensemble member for an action sequence $\vu$. 
The auto-regressive prediction follows a recursive relation:
\begin{equation*}
    [\bar x_{t+1}, \Sigma_{t+1}] = \gp(\bar x_t, u_t)
\end{equation*}
We make use of this in order to estimate the epistemic uncertainty $\ue$.
At each time point of the predicted sequence of observations, we take the empirical variance of the outputted Gaussian parameters $\gp(\bar x_t, u_t)$, predicted from the previous mean prediction $\bar x_t$ and control $u_t$, across the ensembles for that time slice in the predicted trajectories.
This is then summed up across horizon $H$ to obtain the epistemic bonus for action sequence $\vu$.

\begin{figure}
    \centering
 \includegraphics[width=0.5\textwidth]{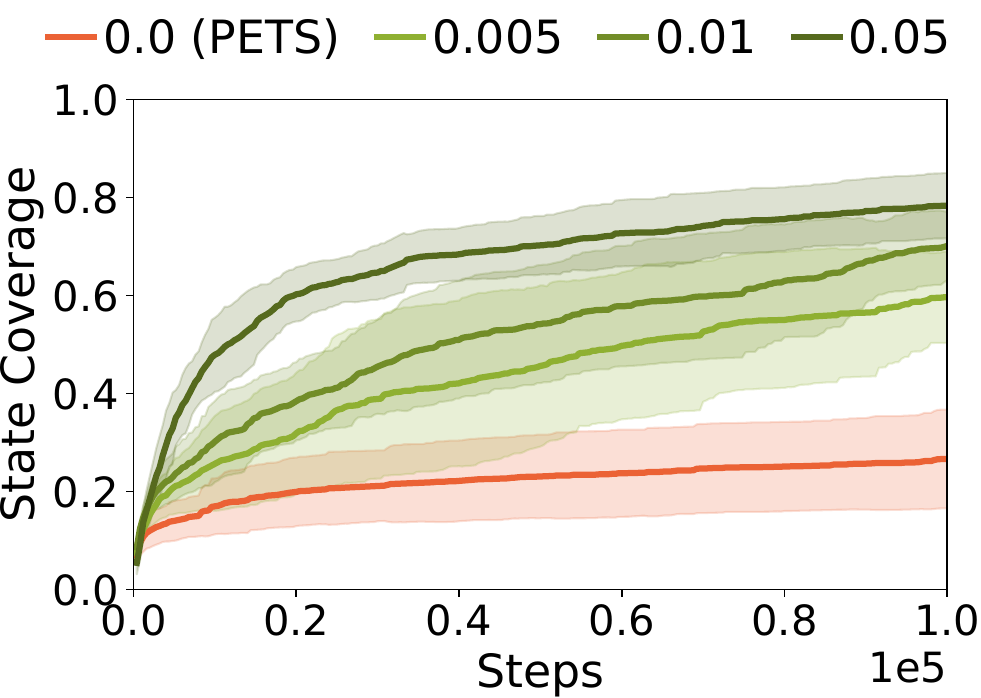}\vspace{-.3em}
     \caption{Exploration over time.}
     \label{suppl_fig:exploration_over_time}

\end{figure}

Fig.\ \ref{suppl_fig:exploration_over_time} shows that scaling $w_\ue$ results in better state-coverage.
This is of particular interest if we want to learn models that are able to generalize to different task settings, \eg when changing the cost function.
While the naive PETS algorithm overfits the model to the task at hand, RAZER learns a truly task-agnostic model and is able to reap the benefits of model-based approaches to control.

For the aleatoric penalty we rely on the actual predictions of the covariance $\Sigma(x_t, u_t)$ and average them across the time slice, following with the sum across horizon $H$. Alternatively to this, we also use the entropy of the Gaussian as the $\ua$ uncertainty measurement.
In Sec. \ref{supp_sec:entropy_vs_variance} we argue how these terms are interchangeable.

 Note that, for the safety term ideally we want to use the full distribution $\psi_\tau$ and separation in aleatoric and epistemic uncertainty is neither required nor desirable.  
 
 \subsection{Entropy vs. Variance as Uncertainty Measurement}\label{supp_sec:entropy_vs_variance}
 
 We use entropy of Gaussian and variance interchangebly as uncertainty estimates.
 Indeed, since the Gaussian distribution is the maximum entropy distribution for certain variance $\sigma^2$, the entropy scales linearly with $\log \sigma^2$.
 We have found that utilizing the variance directly causes RAZER to be much more risk-averse, which can be explained by the variance not being suppressed by the $\log$ term in the entropy.
 Moreover, using the variance directly is much more interpretable and easier to tune because it's of the same scale as the observation space.

\subsection{Observation Space vs. Cost Space Uncertainty}

A natural question to ask when attempting to make efficient use of uncertainties in MPC is where to measure these uncertainties.
As an alternative to observation space uncertainties, one could measure uncertainty in cost space.
Here we argue why this is not a reasonable thing to do for each of the individual cost terms.

\paragraph{Epistemic Bonus} Since we operate under the desiderata that the benefit of model-based methods is in task-agnosticism, we shouldn't measure epistemic uncertainty in the cost space, since this would decouple the task definition through the cost from the observation space and would lead to learning models that are not task-agnostic.

\paragraph{Aleatoric Penalty} This is perhaps the most questionable case for using observation space uncertainty instead of cost space uncertainty.
Nevertheless, we assume that high-aleatoric uncertainty translates to control difficulty, and we want to avoid parts of the observation space that are difficult to control. 
Moreover, the uncertainty measurements become completely invalidated in the case of a task switch, which plays against the task-agnosticism desiderata.

\paragraph{Safety Penalty} Safety is something that is enforced by infusing the algorithm with prior knowledge through a set of constraints which mostly manifest themselves as subsets of the observation space $\gX$ or action space $\gU$. 

\section{Algorithm}\label{supp_sec:algo}

In Algo. \ref{algo:razer} we provide an overview of the CEM algorithm that we utilize for implementing RAZER.
Concretely, we use an improved sample efficient version of CEM as proposed by \citet{pinneri2020icem} that involves shift-initialization of the distribution mean, sampling time-correlated noise and further improvements.

\begin{algorithm}
  Parameters:

  \quad $N$: number of samples; $B$: Number of particles, $H$: planning horizon; $w_{\ua}$, $w_{\ue}$, $w_{\sav}$ CEM-iterations

  \For{t = 1 \textbf{to} T \tcp*{loop over episode length}}{

    \For{i = 1 \textbf{to} CEM-iterations}{

      $(\mathrm{samples}_{p})_{p=1}^{P}$ $\leftarrow$ $N$ samples from CEM$(\mu_t^i, \Sigma_t^i)$, with $P$ particles per sample

      $c$, $c_{\ua}$, $c_{\ue}$, $c_{\sav}$ $\leftarrow$ compute cost functions over particles
      
      $c_{\mathrm{tot}} = c + c_{\ua} + c_{\ue} + c_{\sav}$  \tcp*{compute total cost}

      elite-set$_t$ $\leftarrow$ best $K$ samples according to total cost

      $\mu_t^{i+1}$, $\Sigma_t^{i+1}$ $\leftarrow$ fit Gaussian distribution to elite-set$_t$
    }

    execute first action of best elite sequence 
    
    shift-initialize $\mu_{t+1}^1$
  }
  \caption{RAZER: Risk-aware and safe CEM-MPC}
 \label{algo:razer}
\end{algorithm}

\section{Environments}\label{supp_sec:envs}

All environments are based on the MuJoCo physics engine~\cite{todorov2012mujoco}.
The \textbf{Noisy-Halfcheetah} and \textbf{Noisy-FetchPickAndPlace} environments are based on \textit{HalfCheetah-v3} and \textit{FetchPickAndPlace-v1}, respectively.

\paragraph{BridgeMaze} We designed the \textit{BridgeMaze} environment to show the different aspects of uncertainty, namely the epistemic and aleatoric uncertainty, in isolation. The agent is a simple cube with only a free joint attached to it. The state-space $x=[x_{0}, x_{1}, x_{2}, a, b, c, d, v_{x_{0}}, v_{x_{1}}, v_{x_{2}}]$ is $10$-dimensional, consisting of $3$ positional ($x_{0}$ to $x_{2}$), $4$ rotational ($a$ to $d$) and $3$ velocity-based ($v_{x_{0}}$ to $v_{x_{2}}$), agent-centric coordinates. The action-space $u=[\tau_{x_{0}}, \tau_{x_{1}}]$ is $2$-dimensional. The torque $\tau$ applied to the agent in $x_{0}$- and $x_{1}$-direction.

The task in the environment is to reach a goal platform at $x_{0}^{\star} \geq 12$ by crossing one of three bridges that go over deadly lava.

The domain reward is defined as
\begin{equation}
    r_{t}(x_{t},u_{t},x_{t+1}) = 
        \begin{dcases*}
            | (x_{0})_{t} - x_{0}^{\star} | - | (x_{0})_{t+1} - x_{0}^{\star} | &, if $(x_{1})_{t+1} \geq -1.5$ \\
            0 &, if $(x_{0})_{t+1} \geq x_{0}^{\star}$ and $(x_{1})_{t+1} \geq -1.5$\\ 
            -1 &, otherwise
        \end{dcases*}
\end{equation}

where $x^{\star}$ is the goal state. We define the cost for planning as $c_{t}(x_{t},u_{t},x_{t+1}) = - r_{t}(s_{t},u_{t},s_{t+1})$.

We designed the environments such that the agent is able to accelerate fast and also comes to a full stop relatively fast if no torque is applied. This makes the control problem and the task of learning the model relatively easy.

Noise is added in form of an external force in $x_{1}$-direction injected through the \texttt{xfrc\_applied} attribute of the model. The sign of the force, as well as the force amplitude, sampled from $f_{\textrm{ext}} \in \mathcal{U}(0, f_{\textrm{ext}}^{\textrm{max}})$, are randomly changing every $5$ simulation steps. The external force is added only if $-8 \leq x_{0} \leq 8$ and $-3.6 \leq x_{1} \leq 3.6$. Otherwise the external force is zero.

\paragraph{Noisy-HalfCheetah} We utilize a modified HalfCheetah environment where we apply a normally distributed noise term $\xi \sim \gN(\vmu, \Sigma)$ to the simulator state in the case when the velocity of the cheetah is greater than 6.
More concretely, let $s_t$ denote the simulator state at time step $t$, then the modified state is calculated as follows:
\begin{equation}
   s_t'= s_t + \xi_t
\end{equation}
In our case, $\Sigma$ is a diagonal covariance matrix with the diagonal terms equal to $0.2$. In addition, for the safety experiments with the Noisy-HalfCheetah we create a virtual ceiling at height $h=0.3$. In the case that the body height crosses this threshold, the agent incurs a large penalty.
When the safety-constraint is violated, we don't end the episode.

\paragraph{Noisy-FetchPickAndPlace}
We  modified the \textit{FetchPickAndPlace-v1} environment to show the effect of the aleatoric penalty on the CEM action plan. Given the difficulty of the task, we performed the experiments without the learned model, using instead an ensemble of noisy ground truth dynamics. In this way, we could more easily understand the role of the aleatoric uncertainty during planning.

The noise term  $\xi \sim \gN(\vmu, \Sigma)$  is applied to the action controlling the gripper state: a positive additive noise forces the robot to open the grip with a force proportional to the noise magnitude. This noise is applied to all the ground truth models of the ensemble, and to the environment as well.

In particular, the box position is centered at y-coordinate -1.5 while the target is at $y=2.0$. The gripper state is noisy until $y=1.67$, right before the target.

\paragraph{Solo8-LeanOverObject} The state space of the this environment is $47$-dimensional. It contains the absolute position, rotation, velocity and angular velocity of the robot as well as the positions and velocities of all the joints. In addition, the state contains the positions of the end-effectors and of the sites at the front and back of the robot. The actions space is $8$-dimensional and controls the relative position of the joints. We fixed the two front legs of the robot with a soft-constraint to the ground to prevent the robot from uncontrollable jumping. 
We apply Gaussian noise to the action with a mean of $0$ and a diagonal covariance matrix with the diagonal elements all being $0.3$. The noise is uniformly applied over the entire state-action-space.

The experiments for the \textit{Solo8-LeanOverObject} environment use the ground truth model during planning. The same noise were applied in the 'mental' as well as the 'real' environment.

\subsection{Computing State-Space Coverage}

For computing the state coverage in \fig{fig:exploration_over_time} we divided the continues state-space in $50$ equally spaced bins in the range $-20 \leq x_{0} \leq 20$ and $-10 \leq x_{1} \leq 15$. The state space-coverage is the fractions between states visited at least once and the total number of states.

\section{Application to Transfer Learning}\label{supp_sec:transfer_learning}

In this work we have demonstrated that an approach such as PETS\cite{chua2018:pets} to data-driven MPC that relies on zero-order trajectory optimization of the expected cost is not enough to manage uncertain environments and safety constraints.
These problems need to be addressed when dealing with sim-to-real. 
The separation of uncertainties allows us to effectively manage epistemic uncertainty in the real system, which is important for improving the model once distribution shift to the real system happens.
This can be done in a way of combining the epistemic bonus and probabilistic safety constraints, such that the policy explores parts of the state space where there is knowledge to be obtained while avoiding high-cost regions as a consequence of the incurred safety and aleatoric penalties.

In comparison to standard approaches for sim-to-real which involve domain randomization at training time, this approach incurs lower computational overhead and relies on learning on the real system.

\end{document}